\documentclass[runningheads]{llncs}
\usepackage{makecell}
\usepackage[T1]{fontenc}
\usepackage{graphicx,verbatim}
\usepackage{cite}
\usepackage{amsmath,amssymb,amsfonts}
\usepackage{algorithmic}
\usepackage{graphicx}
\usepackage{multirow}
\usepackage{multicol}
\usepackage{mathtools}
\usepackage{mathrsfs}
\usepackage{marvosym}
\usepackage{subfig}
\usepackage{algorithmic}
\usepackage{textcomp}
\usepackage{graphics}
\usepackage{threeparttable}
\usepackage{amsfonts}
\usepackage{float}
\usepackage{bm}
\usepackage{array}
\usepackage{colortbl}
\usepackage{pifont}
\usepackage{diagbox}
\usepackage{rotating}
\usepackage{booktabs}
\usepackage{overpic}
\usepackage{contour}
\usepackage[table]{xcolor}
\usepackage{tikz}
\usepackage{hyperref}

\newcommand{\tabref}[1]{Table \ref{#1}}
\newcommand{\figref}[1]{Fig. \ref{#1}}
\newcommand{\secref}[1]{Sec. \ref{#1}}

\newcommand{\ie}{{\emph{i.e.}}}
\newcommand{\eg}{{\emph{e.g.}}}
\newcommand{\etal}{{\emph{et al.}}}

\def\ourmodel{TopoNet}

\def\miccaimodel{D$^2$GPLand}

\begin{document}
\title{Topology-Constrained Learning for Efficient Laparoscopic Liver Landmark Detection}
%
%
\titlerunning{TopoNet for Efficient Liver Landmark Detection}
%
\author{Ruize Cui\inst{1} \and
Jiaan Zhang\inst{2} \and
Jialun Pei\inst{3(\textrm{\Letter})} \and
Kai Wang\inst{4} \and
Pheng-Ann Heng\inst{3}\and 
Jing Qin\inst{1}}
\authorrunning{R. Cui et al.}
%
\institute{The Hong Kong Polytechnic University, Hong Kong, China \and
National University of Singapore, Singapore \and
The Chinese University of Hong Kong, Hong Kong, China  \and  
Nanfang Hospital, Southern  Medical University, Guangzhou, China \\
\email{jialunpei@cuhk.edu.hk}}



\maketitle              
\begin{abstract}
Liver landmarks provide crucial anatomical guidance to the surgeon during laparoscopic liver surgery to minimize surgical risk. 
%
%
However, the tubular structural properties of landmarks and dynamic intraoperative deformations pose significant challenges for automatic landmark detection. 
%
In this study, we introduce~\ourmodel, a novel topology-constrained learning framework for laparoscopic liver landmark detection. 
%
%
%
Our framework adopts a snake-CNN dual-path encoder to simultaneously capture detailed RGB texture information and depth-informed topological structures. 
Meanwhile, we propose a boundary-aware topology fusion (BTF) module, which adaptively merges RGB-D features to enhance edge perception while preserving global topology.
Additionally, a topological constraint loss function is embedded, which contains a center-line constraint loss and a topological persistence loss to ensure homotopy equivalence between predictions and labels.
%
Extensive experiments on L3D and P2ILF datasets demonstrate that~\ourmodel~achieves outstanding accuracy and computational complexity, highlighting the potential for clinical applications in laparoscopic liver surgery.  Our code will be available at \url{https://github.com/cuiruize/TopoNet}.


\end{abstract}
\section{Introduction}
Laparoscopic liver surgery has been widely adopted due to its perioperative advantages of reduced blood loss, faster recovery, and lower complication rates~\cite{fretland2018laparoscopic,giannone2024robotic}.
%
However, the limited laparoscopic field of view and intraoperative deformations make the identification of key liver anatomical structures particularly challenging.
%
Augmented reality (AR) navigation technology has emerged as a promising solution to provide visual guidance to surgeons by establishing correspondences between intraoperative 2D keyframes and preoperative 3D anatomy~\cite{p2ilf,hu2024artificial}.
%
%
In this regard, accurate intraoperative liver landmark detection is essential to perform high-quality registration with preoperative 3D models.
%
However, automatic intraoperative liver landmark detection remains challenging due to liver deformation, varying laparoscopic viewpoints, and occlusions from outliers like surgical instruments.
As a result, there is still demands for the development of intelligent computer-assisted techniques for robust liver landmark detection in a complex laparoscopic environment.

\begin{figure}[t!]
\centering
\includegraphics[width=\textwidth]{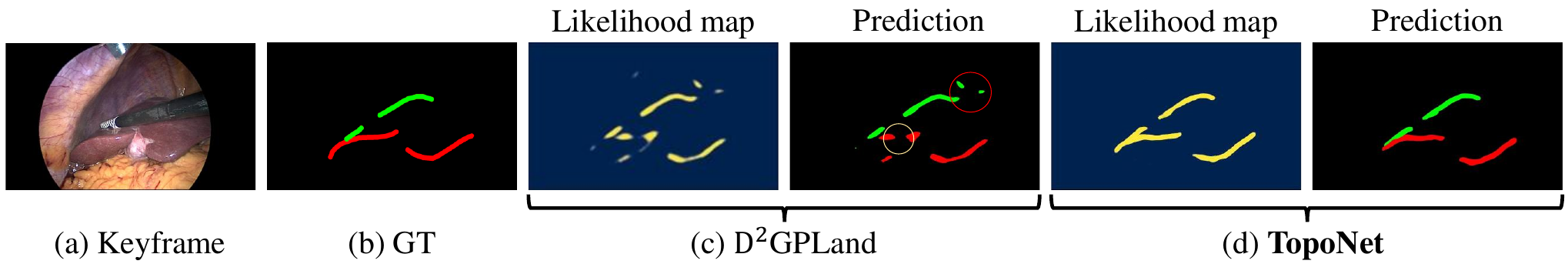}
\caption{Illustration of the crucial role of topological priors in tubular characteristics learning and topology preservation. We highlight a broken landmark in yellow circle and two topological false positives caused by outliers in red circle.}
\label{fig1}
\vspace{-8pt}
\end{figure}
Among various marked forms of liver landmarks, continuous semantic regions provide richer semantic features compared to other forms such as contours~\cite{collins2020augmented} or bounding boxes~\cite{smithmaitrie2024development}, which have been confirmed as valuable for enhancing intraoperative spatial relationships~\cite{schneider2021performance}.
Several existing studies have explored using deep-learning-based models to automatically segment landmark regions.
Labrunie~\etal~\cite{labrunie2022automatic} employed an original U-Net~\cite{u-net} to detect 2D landmarks and then aligned them with preoperative 3D models. 
Subsequently, more advanced networks~\cite{p2ilf}, \eg, nnU-Net~\cite{nnunet} and UNet++~\cite{unet++}, have been applied for more accurate landmark detection.
However, these methods primarily utilize existing segmentation frameworks while overlooking the anatomical characteristics of liver landmarks.
More recently, D$^2$GPLand~\cite{d2gpland} integrates depth geometric priors and prompt-guided training to enhance the learning of liver landmark features, further improving the detection results.
Despite improvements in detection performance, the fine-tubular structural properties of liver landmarks and the presence of outliers still seriously affect the accuracy of landmark detection, as shown in~\figref{fig1} (c).
Most existing approaches concentrate on pixel-level textural and geometric clues, ignoring the intrinsic topology characteristics of liver landmarks, which are effective in mitigating false positive detections from outlier occlusions and preserving the structural continuity of tubular landmarks.
To this end, we consider embedding topological constraints into the network to develop an efficient and precise laparoscopic liver landmark detection algorithm.

In this work, we introduce a topology-constrained learning network, termed \emph{\ourmodel}, for efficient laparoscopic liver landmark detection.
%
Specifically, considering that depth modality has been demonstrated to provide effective auxiliary geometric information~\cite{d2gpland}, we design a \textbf{snake-CNN dual-path encoder}, which employs snake topology acquisition (STA) blocks and a CNN to extract depth topological structures and detail texture features, respectively.
Then, a \textbf{boundary-aware topological fusion (BTF)} module is proposed for adaptively merging RGB-D features and sensing boundary features, which facilitates the preservation of the global topology of liver landmarks (see the likelihood maps in~\figref{fig1}).
%
To adequately exploit topology characteristics for detection supervision, we also present a \textbf{topological constraint loss function} composed of a center-line constraint loss and a topological persistence loss to ensure homotopy equivalence between predictions and labels.
Experimental results on L3D and P2ILF datasets~\cite{d2gpland,p2ilf} show that~\ourmodel~outperforms 12 advanced models on the liver landmark detection task.

\section{Methodology}

\begin{figure}[t!]
\includegraphics[width=\textwidth]{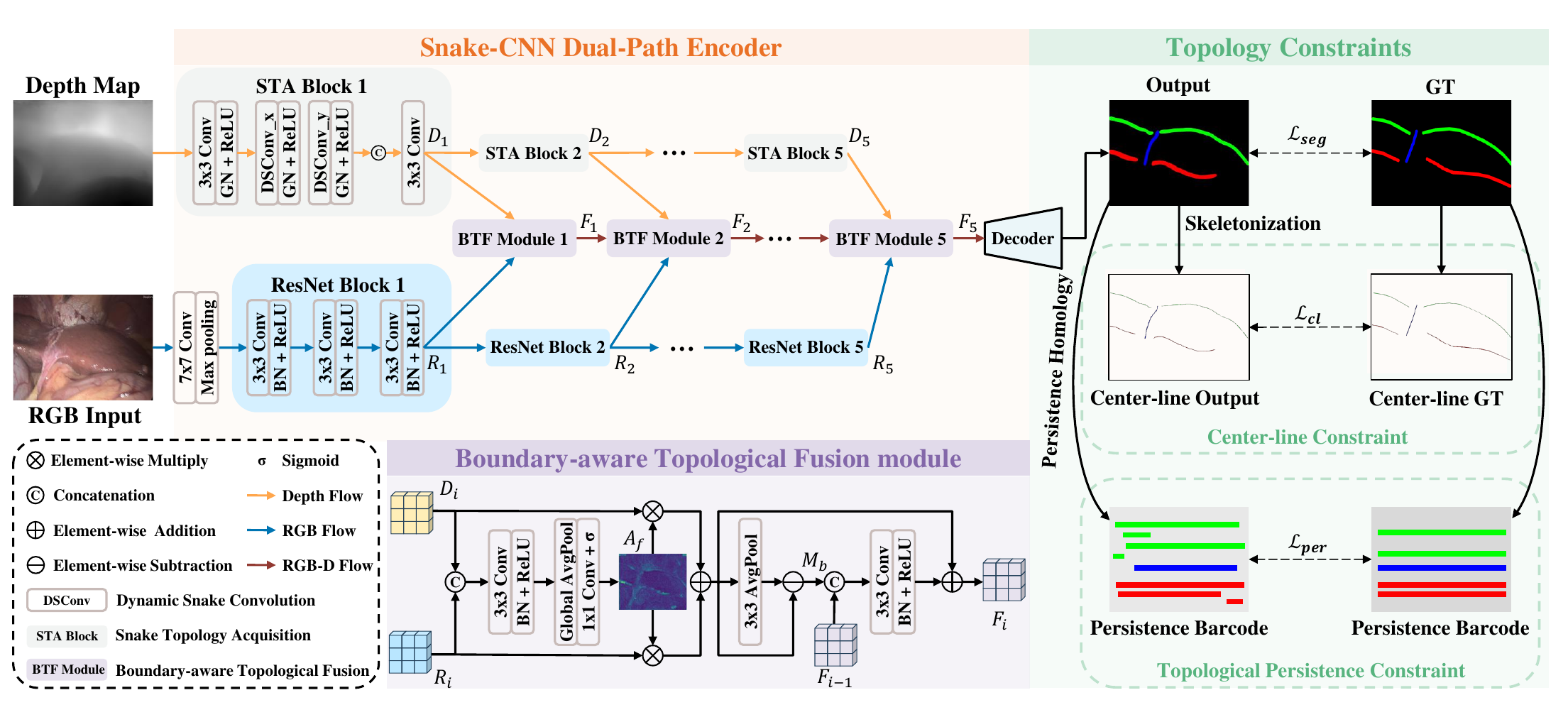}
\caption{Overall architecture of the proposed~\ourmodel. 
}
\label{overview}
\vspace{-5pt}
\end{figure}

The detailed pipeline of~\ourmodel~is illustrated in~\figref{overview}. Our framework takes RGB keyframes and the corresponding depth maps estimated by the frozen AdelaiDepth~\cite{depth} as inputs.
%
We embrace multi-view snake convolutions into the snake topology acquisition (STA) block to capture deep topological cues and combine ResNet~\cite{resnet} to extract multiscale RGB features.
%
Then, the depth feature $D_i$ and RGB feature $R_i$ output from corresponding encoder blocks are passed into the boundary-aware topological fusion (BTF) module to generate the fused features $F_i$.
%
Notably, the first BTF module takes $D_1$ and $R_1$ as inputs, and the other ones utilize only the former fused feature $F_{i-1}$ for residual learning to reduce information loss.
%
Finally, we adopt a CNN decoder to produce landmark results and supervise our network with the proposed topological constraint loss to constrain the connectivity and topological persistence of detection maps.
%

\subsection{Snake-CNN Dual-Path Encoder} 
To perceive semantic landmarks with slender and tortuous tubular structures, we design a snake-CNN dual-path encoder that separately extracts depth topology information and RGB texture clues with our STA blocks and CNN blocks.
%
%
%
%
As illustrated in~\figref{overview}, we employ a ResNet-34 encoder~\cite{resnet} as the backbone for RGB keyframes, \ie, the $i$-${th}$ ResNet block outputs the RGB feature $R_i$.
The depth pathway consists of five cascaded STA blocks for perceiving topological properties.
In $i$-${th}$ STA block, the input depth feature $D_{i-1}$ is first passed through a convolution block for lower-level feature acquisition.
Then, we apply multi-view dynamic snake convolution (DSConv)~\cite{qi2023dynamic} blocks to extract critical topology cues in the X- and Y-axes.
%
Lastly, we concatenate single-axis features and use another convolution block for cross-axis feature aggregation to generate the output depth feature $D_i$.

\subsection{Boundary-aware Topological Fusion} 
The proposed BTF module aims to integrate RGB and depth features effectively to capture comprehensive landmark features and preserve topological structures.
%
As shown in~\figref{overview}, the input dual-modal features $R_i$ and $D_i$ are first concatenated along the channel dimensions and fed into the convolution block to obtain merged features.
%
Then, a global average pooling operation and Sigmoid activation are applied to generate the fused attention map $A_f$ that highlights the important regions.
The fusion process can be formulated as
\begin{equation}
    A_f = \mathbf{\sigma}\big(\texttt{Pg}(\texttt{ReLU}(\texttt{BN}(\texttt{Conv}(\mathcal{C}[{D_i},{R_i}]))))\big),
\end{equation}
where $\texttt{Conv}$ represents the convolution operation, $\texttt{BN}$ is batch normalization. $\texttt{Pg}$ is the global average pooling operation.
Afterward, we multiply $A_f$ by $R_i, D_i$ and element-wise sum them to get the primary fused feature $\hat{F_i}$:
\begin{equation}
    \hat{F_i} = (R_i \cdot A_f)~\oplus~(D_i \cdot A_f),
\end{equation}
where $\oplus$ stands for the element-wise summation.
Upon getting $\hat{F_i}$, we design a boundary enhancement operation to guide the model to focus on the ambiguous edge regions.
Here the principle is that liver landmarks are ultimately the boundary areas of the liver, and focusing on these regions is beneficial for holistic learning of anatomical structures, thus preserving the global topology of liver landmarks, \emph{i.e.}, the relationships among different types of landmarks.
Concretely, we apply an average pooling with $3\times3$ kernel and compute the feature differences by element-wise subtraction to obtain the boundary map $M_b$.
%
After that, we perform multi-scale aggregation of $M_b$ and the output of the previous BTF module $F_{i-1}$ with convolution operation and add $\hat{F_i}$ in the form of residual to reduce the information loss, resulting in the output $F_i$:
\begin{equation}
    F_i = \hat{F_i}~\oplus~\texttt{Conv}\big(\texttt{BN}(\texttt{ReLU}(\mathcal{C}[(\hat{F_i}-\texttt{Po}(\hat{F_i})),F_{i-1}]))\big),
\end{equation}
where $\texttt{Po}$ indicates the average pooling operation.

\subsection{Topological Constraint Loss}\label{topology constraint}
To further refine the topological structure, we introduce the topological constraint loss function for supervision, involving a center-line constraint loss and a topological persistence loss.
%

\subsubsection{Center-line Constraint Loss.}
Since tubular structures rely heavily on connectivity, traditional pixel-wise segmentation losses (\eg, Dice Loss) that mainly focus on overlapping areas usually overlook structural discontinuities.
%
To this end, we inject the center-line supervision into the loss function for ensuring landmark continuity. This supervision extends the clDice loss function~\cite{cldice} to a multi-class form to supervise our network.
%
%
Given the prediction $P_l$ of landmark class $l$ and its label $G_l$, we can extract their skeletons $S_p^l$ and $S_g^l$. The topology precision $T_{prec}$ and topology sensitivity $T_{sens}$ can be defined as
\begin{equation}
    T_{prec}(S_p^l, G_l) = \frac{|S_p^l \cap G_l|}{|S_p^l|}, \ T_{sens}(S_g^l, P_l) = \frac{|S_g^l \cap P_l|}{|S_g^l|}.
\end{equation}
Then we can compute the multi-class center-line constrained loss $\mathcal{L}_{cl}$:
\begin{equation}
    \mathcal{L}_{cl} = \frac{1}{L}\sum_{l=1}^L{\big(2\times\frac{T_{prec}(S_p^l, G_l)\times{T_{sens}(S_g^l, P_l)}}{T_{prec}(S_p^l, G_l)+T_{sens}(S_g^l, P_l)}\big)},
\end{equation}
where $L=3$ denotes the number of landmark classes to be detected.

\subsubsection{Topological Persistence Loss.}
%
Impacted by outliers present in laparoscopic scenes, including instruments, blood, and similarly textured tissues, the network is prone to produce topological false positive results.
%
To maintain the topological consistency between predictions and labels, we introduce a novel topological persistence loss $\mathcal{L}_{per}$ based on the persistent homology theory~\cite{persistence_homology}.
%
%
Given a predictive likelihood map $Y_l$ of landmark class $l$ and its ground truth $G_l$, we exploit the efficient barcode computation algorithm~\cite{bettimatch} to obtain the persistence barcodes of $Y_l$ and $G_l$ in order to obtain the sets of birth and death coordinates of the matched connected components (denoted by $B^m$ and $D^m$) and the unmatched ones (denoted by $B^u$ and $D^u$).
For each matched connected component $C$, we can find the persistence interval $C_{pre} = (B_{pc}^m, D_{pc}^m)$ in prediction from $B^m$ and $D^m$, while the persistence interval in GT is denoted by $C_{gt} = (B_{gc}^m, D_{gc}^m)$.
Here, we can obtain the persistence loss for the matched connected components by calculating the difference of the persistence intervals of $C$ in prediction and GT:
\begin{equation}
    \mathcal{L}_m^l = \frac{1}{N_m^l + s}\cdot\sum_{i\in M}~(||B_{pc}^m - B_{gc}^m||_2 + ||D_{pc}^m - D_{gc}^m||_2),
\end{equation}
where $N_m^l$ is the number of matched connected components of class $l$ and $s = 1e^{-5}$ denotes the smoothing factor. M is the set of matched components.
For each unmatched connected component $Z$ in prediction, we can also get its persistence interval by $Z_{pre} = (B_{pz}^u, D_{pz}^u)$.
Then we can compute the persistence loss for the unmatched components by calculating the length of its persistence interval:
\begin{equation}
    \mathcal{L}_u^l = \frac{1}{N_u^l + s}\cdot\sum_{i\in U}~||B_{pz}^u - D_{pz}^u||_2,
\end{equation}
where $N_u^l$ denotes the number of unmatched connected components of $l$ and $U$ is the set of unmatched components in prediction. 
The final $L_{per}$ is derived by summing the individual losses and expanding to all landmark classes:
\begin{equation}
    \mathcal{L}_{per} = \frac{1}{L}\sum_{l=1}^L ~(\frac{N_m^l}{N_m^l+N_u^l}\cdot\mathcal{L}_m^l + \frac{N_u^l}{N_m^l+N_u^l}\cdot\mathcal{L}_u^l).
\end{equation}
Lastly, we combine two topological constraint losses with Dice loss $\mathcal{L}_{dice}$ to reach the total loss function:
\begin{equation}
    \mathcal{L}_{total} = \lambda_d \cdot \mathcal{L}_{dice} + \lambda_{cl} \cdot \mathcal{L}_{cl} + \lambda_{per} \cdot \mathcal{L}_{per},
\end{equation}
where $\lambda_d, \lambda_{cl}, \lambda_{per}$ are the balancing parameters for each loss function.

\section{Experiments}

\subsection{Datasets and Metrics}
We conduct experiments on two laparoscopic liver landmark datasets: L3D~\cite{d2gpland} and P2ILF~\cite{p2ilf}.
L3D contains three liver landmark categories and consists of 1,152 annotated keyframes of 1920$\times$1080 pixels from liver surgical videos. 
%
P2ILF has 183 annotated laparoscopic images with the same landmark categories as L3D for liver landmark detection.
Wherein, 167 images are used for training while the others are used for testing.
Since only the training set can be available, we randomly select 124 images for training and the remaining 43 images for testing.

For evaluation metrics, we follow the experimental setup in~\cite{d2gpland}, utilizing the Intersection over Union (IoU), Dice Score Coefficient (DSC), and Average Symmetric Surface Distance (Assd). 
We also compute the average inference speed and GFLOPs to evaluate the model efficiency.

\begin{table}[t]
    \renewcommand{\arraystretch}{1.2}
    \centering
    \caption{Comparison with cutting-edge methods on L3D and P2ILF test sets.}
    \scriptsize
    \resizebox{\linewidth}{!}{
    \begin{tabular}{l|ccc|ccc|c|c}
        \hline
        \multirow{2}{*}{Methods} & \multicolumn{3}{c|}{L3D~\cite{d2gpland}}& \multicolumn{3}{c|}{P2ILF~\cite{p2ilf}}& \multirow{2}{*}{Infer. Speed~$\downarrow$}  & \multirow{2}{*}{GFLOPs~$\downarrow$}\\
        \cline{2-7}
        & DSC~$\uparrow$ & IoU~$\uparrow$ & Assd~$\downarrow$ & DSC~$\uparrow$ & IoU~$\uparrow$ & Assd~$\downarrow$ && \\ \hline
        U-Net \cite{u-net} & 51.39 & 36.35 & 84.94 & 29.89 & 17.89 & 47.95 &172.65ms& 774.30 \\ 
        COSNet \cite{labrunie2023automatic} & 56.24 & 40.98 & 69.22 & 26.78 & 16.05 & 47.33 &167.25ms & 319.24\\ 
        Res-UNet \cite{resunet} & 55.47 & 40.68 & 70.66 & 25.84 & 15.76 & 50.97 &155.76ms& 314.65\\ 
        UNet++ \cite{unet++} & 57.09 & 41.92 & 74.31 & 34.96 & 21.57 & 42.39 &213.19ms& 554.87 \\ 
        HRNet \cite{hrnet} & 58.36 & 43.50 & 70.02 & 33.64 & 21.31 &44.16& 146.82ms & 376.52 \\ 
        TransUNet \cite{transunet} & 56.81 & 41.44 & 76.16 & 25.49 & 15.22 & 52.14 &262.21ms& 795.10 \\
        Swin-UNet \cite{swinunet} & 57.35 & 42.09 & 72.80 & 18.65 & 10.62 & 68.46 & \underline{127.22ms} & \textbf{129.68} \\ 
        SAM-Adapter \cite{samada} & 57.57 & 42.88 & 74.31 & 21.12 & 12.00 & 57.13 &300.21ms& 489.35 \\ 
        SAMed \cite{samed} & 62.03 & 47.17 & 61.55 & 31.73 & 19.42 & 40.08 & 278.56ms & 488.95 \\ 
        SAM-LST \cite{samlft} & 60.51 & 45.03 & 68.87 & 28.75 & 18.38 &46.53& 304.78ms & 517.68 \\                 
        AutoSAM \cite{autosam} & 59.12 & 44.21 &62.49& 25.71 & 15.43 & 53.65 &337.44ms& 589.21 \\ 
        \miccaimodel \cite{d2gpland} & \underline{63.52} & \underline{48.68} & \underline{59.38} & \underline{40.55} & \underline{25.87} & \underline{38.73} & 297.93ms & 572.85\\ \hline
        \textbf{\ourmodel~(Ours)} & \textbf{65.19} & \textbf{50.56} & \textbf{28.07} & \textbf{41.36} & \textbf{26.88}& \textbf{30.16}& \textbf{86.43ms} &  \underline{276.99} \\ \hline
    \end{tabular}
    }
    \label{results}
\end{table}

\subsection{Implementation Details}\label{experiment}
The training and testing processes of~\ourmodel~are executed on a single NVIDIA RTX A6000 GPU. We train 100 epochs with a batch size of 4. 
We employ the Adam optimizer with the initial learning rate of 8e-5 and weight decay factor of 3e-5. 
In addition, the CosineAnnealingLR scheduler is used to adjust the learning rate to 1e-6 at the end of the training. 
A warmup strategy is also applied to adaptively embed topology constraints. 
We empirically set $\lambda_d = 0.4$, $\lambda_{cl} = 0.4$, and $\lambda_{per} = 0.2$ for balancing parameters. 

\subsection{Comparison with State-of-the-art Methods}
\begin{figure*}[t!]
\centering
\includegraphics[width=\textwidth]{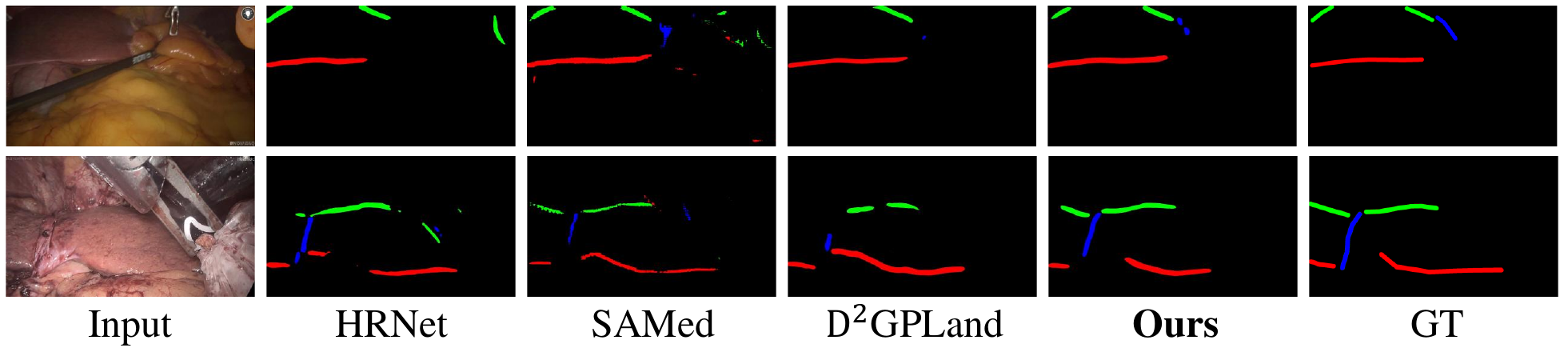}
\caption{Visual comparison of our~\ourmodel~with representative models.} \label{fig3}
\end{figure*}
We follow the experimental settings of L3D benchmark to compare~\ourmodel~with 12 cutting-edge methods. 
%
%
As shown in~\tabref{results},~\ourmodel~achieves more precise detection results on both datasets than other models.
Compared to the second-ranked~\miccaimodel, \ourmodel~improves 1.67\% on DSC, 1.88\% on IoU, and \textbf{31.31} pixels on Assd on L3D dataset.
%
For the P2ILF dataset, our method improves the performance on the DSC, IoU, and Assd metrics by 0.81\%, 1.01\%, and \textbf{8.57} pixels, respectively.
%
Notably, we observe that the performance improvement in the Assd metric is tremendous.
Assd is defined to compute the shortest distance from each foreground pixel in prediction to the foreground in GT.
As mentioned in~\secref{topology constraint}, the outliers in surgical keyframes affect the model to produce topological false positives in prediction, which are not typically located in locations that overlap with the target landmarks, and each of the pixel in these topologically inaccurate predictions contributes significantly to the Assd metric.
The topological constraint loss, especially $L_{per}$, proposed in our method can suppress the negative effects of these outliers and preserve the topology structure, resulting in a substantial improvement in the Assd metric.
%

For model efficiency, the inference speed of~\ourmodel~outperform all compared models and the computational complexity of our model also reaches the leading level.
Compared to~\miccaimodel, the inference speed of~\ourmodel~is about four times faster while requiring only about half the GLOPs.
%
%
%
\figref{fig3}~also exhibits the visualization results of~\ourmodel~and competitors.
By intensive learning of topological characteristics, our method produces more accurate results while alleviating the influences of outliers.

\subsection{Ablation Analysis}\label{abla}
We conduct ablation studies on the key components of our model, including the boundary-aware topological fusion (BTF) module, center-line constraint loss $\mathcal{L}_{cl}$, and topological persistence loss $\mathcal{L}_{per}$.
We use the evaluation set of L3D for experiments.
In our baseline, we replace the BTF module with simple RGB-D concatenation and remove the proposed topological losses.
%
As illustrated in~\tabref{ablation1}, the topological losses contribute to the detection performance, and the addition of $\mathcal{L}_{per}$ brings a significant improvement in Assd.
%
When embedding the BTF module for comprehensive RGB-D fusion, there is also an improvement in detection performance with 1.87\% in DSC, 1.35\% in IoU, and 8.20 pixels in Assd.
%
In short, both our BTF module and topological constraint loss functions play an indispensable role in optimizing the model performance.

\begin{figure*}[t!]
\centering
\includegraphics[width=\textwidth]{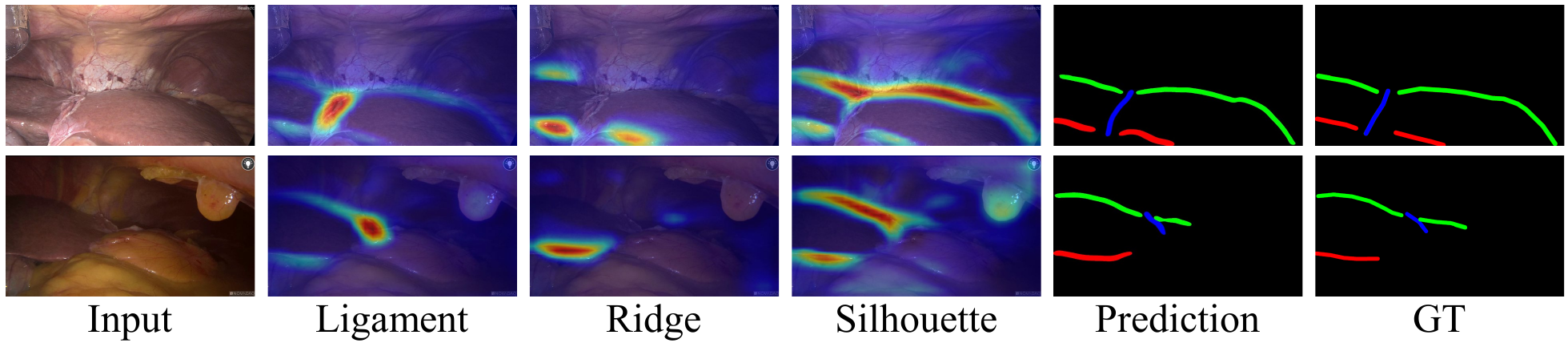}
\caption{Visualizations of class-aware attention maps with our predictions.} \label{fig4}
\end{figure*}

\begin{figure}[!t]
    \begin{minipage}{0.45\textwidth}
        \footnotesize
        \centering

        \captionof{table}{Ablations for key Designs.}
        \vspace{-5pt}
        \scalebox{0.97}{
        \begin{tabular}{c|c|c|c}
        \hline
        Methods & DSC & IoU & Assd\\
        \hline
        Baseline & 54.64& 40.94& 49.64\\ \hline
         w/o $\mathcal{L}_{per}$ & 58.87 & 46.69 & 38.66\\
         w/o $\mathcal{L}_{cl}$ & 58.63 & 46.37 & 32.01\\
         w/o $\mathcal{L}_{per}$ \& $\mathcal{L}_{cl}$ & 57.44 & 45.86 & 43.32\\ \hline
         w/o BTF & 57.92 & 46.03 & 37.47 \\ \hline    
         \rowcolor{black!5}\textbf{\ourmodel} & \textbf{59.79} & \textbf{47.38} & \textbf{29.27} \\  
        \hline
        \end{tabular}
        }
        \label{ablation1}
    \end{minipage}
    \vspace{-3pt}
    \begin{minipage}{0.03\textwidth}
    \begin{tabular}{c}
          
    \end{tabular}
    \end{minipage}
    \begin{minipage}{0.52\textwidth}
        \centering
        \includegraphics[width=0.8\textwidth,height=0.47\textwidth]{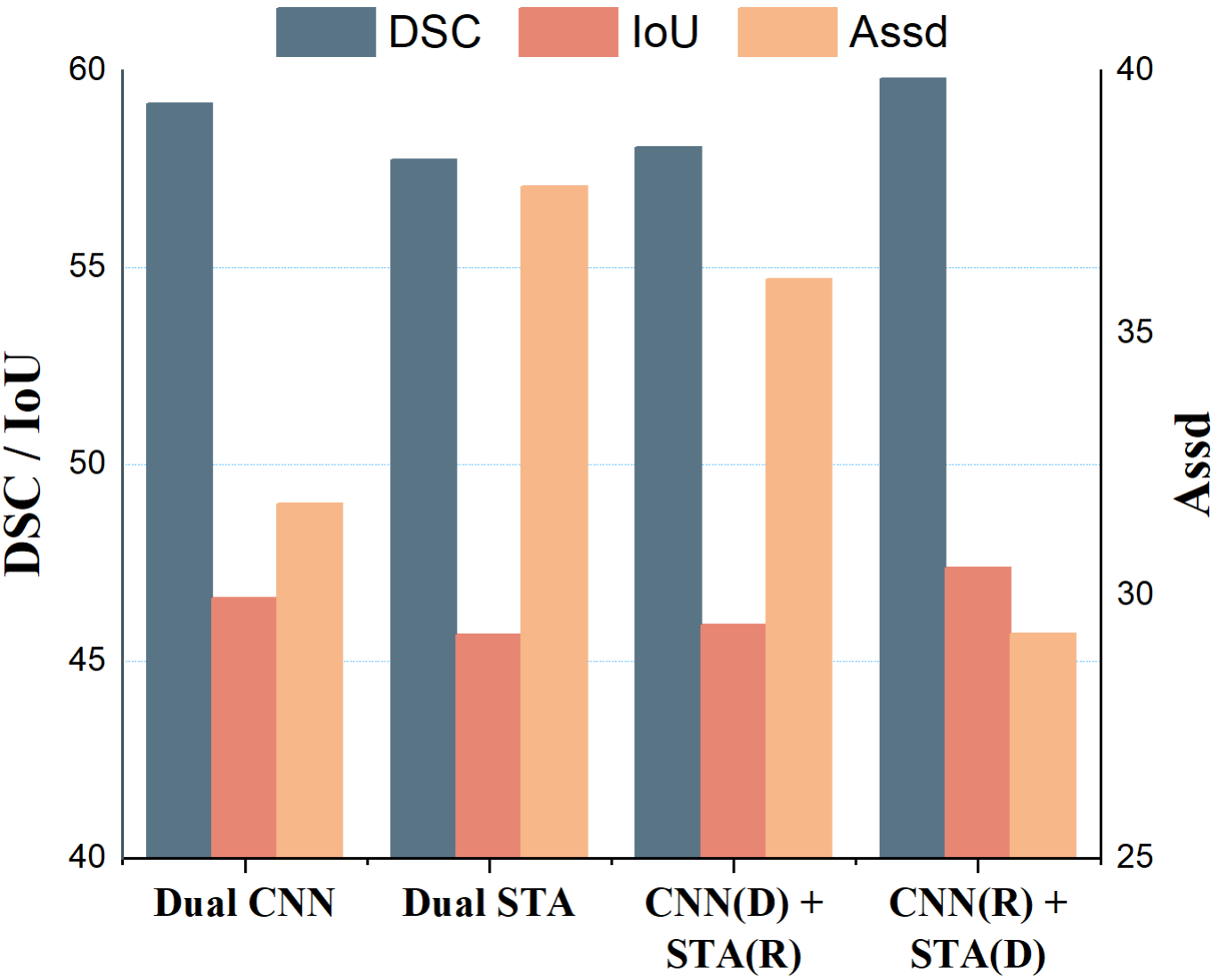}
        \vspace{-6pt}
        \caption{Ablations for backbones.} \label{backbone}
    \end{minipage}
    
\end{figure}

%
We also analyze the effect of different backbones on the snake-CNN encoder in RGB-D feature extraction.
%
As shown in~\figref{backbone}, when applying ResNet-34 blocks for RGB keyframes and the STA blocks for depth maps, our framework achieves optimal performance in all metrics.
%
We also display the attention maps from the BTF Module 5 (refer to~\figref{overview}) for two typical samples in~\figref{fig4}.
%
It can be seen that our~\ourmodel~enables accurate detection of landmarks and precise identification of landmark-related regions from a topological standpoint.

\section{Conclusion}
This study presents a lightweight topological-constrained framework~\ourmodel~for precise and efficient laparoscopic liver landmark detection.
Our framework comprises a snake-CNN dual-path encoder for RGB-D feature extraction, coupled with a boundary-aware fusion module to integrate bi-modal features and preserve the topology.
More importantly, a topological constraint loss is introduced to enhance the learning of topological characteristics of liver landmarks and prevent topological errors.
Our method outperforms current state-of-the-art methods with faster inference.
This work opens up new possibilities for precise and efficient liver landmark detection and facilitates the application in clinical surgery.

\bibliographystyle{splncs04}
\bibliography{ref}

\end{document}